\documentclass[10pt,twocolumn,letterpaper]{article}

\usepackage[pagenumbers]{cvpr} 

\definecolor{cvprblue}{rgb}{0.21,0.49,0.74}

\definecolor{mygray}{gray}{.88}

\usepackage{times}
\usepackage{epsfig}
\usepackage{graphicx}
\usepackage{amsmath}
\usepackage{amssymb}
\usepackage{booktabs}
\usepackage{multirow}
\usepackage{makecell}
\usepackage{comment}
\usepackage[table]{xcolor}
\usepackage[linesnumbered,ruled,vlined]{algorithm2e}
\usepackage{diagbox}
\usepackage{placeins}
\usepackage{colortbl}
\usepackage{bm}
\usepackage{pifont}
\usepackage{subcaption}
\usepackage{tabularx}
\usepackage{threeparttable}
\usepackage{tikz}
\usepackage{microtype}

\SetAlFnt{\small}
\SetKwComment{Comment}{/* }{ */}
\SetKwInput{kwInput}{Input}
\SetKwInput{kwOutput}{Output}
\RestyleAlgo{ruled}

\usepackage{array}
\usepackage[normalem]{ulem}
\useunder{\uline}{\ul}{}
\usepackage{float, wrapfig}
\usepackage{resizegather}
\usepackage{enumitem}
\usepackage{xspace}
\usepackage{url}
\usepackage{adjustbox}

\usepackage[font=small,skip=0pt]{caption}
\usepackage[accsupp]{axessibility}

\newcommand{\std}[1]{{\scriptsize$\pm$#1}}

\usepackage[pagebackref,breaklinks,colorlinks,allcolors=cvprblue]{hyperref}

\usepackage[capitalize]{cleveref}
\crefname{section}{Sec.}{Secs.}
\Crefname{section}{Section}{Sections}
\Crefname{table}{Table}{Tables}
\crefname{table}{Tab.}{Tabs.}

\def\paperID{17212} 
\def\confName{CVPR}
\def\confYear{2026}

\title{Accelerating Large-Scale Dataset Distillation via Exploration–Exploitation Optimization}

\author{
Muhammad J. Alahmadi\textsuperscript{1,2} \hspace{2mm}
Peng Gao\textsuperscript{1} \hspace{2mm}
Feiyi Wang\textsuperscript{3} \hspace{2mm}
Dongkuan (DK) Xu\textsuperscript{1} \\
\textsuperscript{1}North Carolina State University \hspace{4mm}
\textsuperscript{2}King Abdulaziz University \hspace{4mm}
\textsuperscript{3}Oak Ridge National Laboratory \\
{\tt\small \{mjalahma,pgao5,dxu27\}@ncsu.edu} \hspace{4mm}
{\tt\small fwang2@ornl.gov} \hspace{4mm}
{\tt\small mjalahmadi@kau.edu.sa}
}

\begin{document}
\maketitle
\begin{abstract}

Dataset distillation compresses the original data into compact synthetic datasets, reducing training time and storage while retaining model performance, enabling deployment under limited resources. Although recent decoupling-based distillation methods enable dataset distillation at large-scale, they continue to face an efficiency gap: optimization‑based decoupling methods achieve higher accuracy but demand intensive computation, whereas optimization‑free decoupling methods are efficient but sacrifice accuracy. To overcome this trade‑off, we propose Exploration--Exploitation Distillation (E$^2$D), a simple, practical method that minimizes redundant computation through an efficient pipeline that begins with full-image initialization to preserve semantic integrity and feature diversity. It then uses a two‑phase optimization strategy: an exploration phase that performs uniform updates and identifies high‑loss regions, and an exploitation phase that focuses updates on these regions to accelerate convergence. We evaluate E$^2$D on large-scale benchmarks, surpassing the state-of-the-art on ImageNet-1K while being 18$\times$ faster, and on ImageNet-21K, our method substantially improves accuracy while remaining 4.3$\times$ faster. These results demonstrate that targeted, redundancy-reducing updates, rather than brute-force optimization, bridge the gap between accuracy and efficiency in large-scale dataset distillation. Code is available at \url{https://github.com/ncsu-dk-lab/E2D}. 

\end{abstract}    
\section{Introduction}
\label{sec:intro}

Dataset distillation, also known as dataset condensation, aims to distill the original data into compact synthetic datasets that retain the key information of the original data while being orders of magnitude smaller~\cite{tongzhouw2018datasetdistillation,cazenavette2022distillation,Wang_2022_CVPR,zhao2021DC}. Distilled datasets train models faster and require less storage, enabling adaptability and real‑world deployment under tight time and resource budgets~\cite{tongzhouw2018datasetdistillation,zhao2021DSA,song2022federated}. Despite recent advances, dataset distillation remains constrained by high computational cost, with the most common bi-level distillation methods requiring days to synthesize even moderately sized datasets (e.g., CIFAR‑100) ~\cite{zhao2021DC,kimICML22_IDC,Acc-DD}.

\begin{figure}[t]
  \centering
   \includegraphics[width=\linewidth]{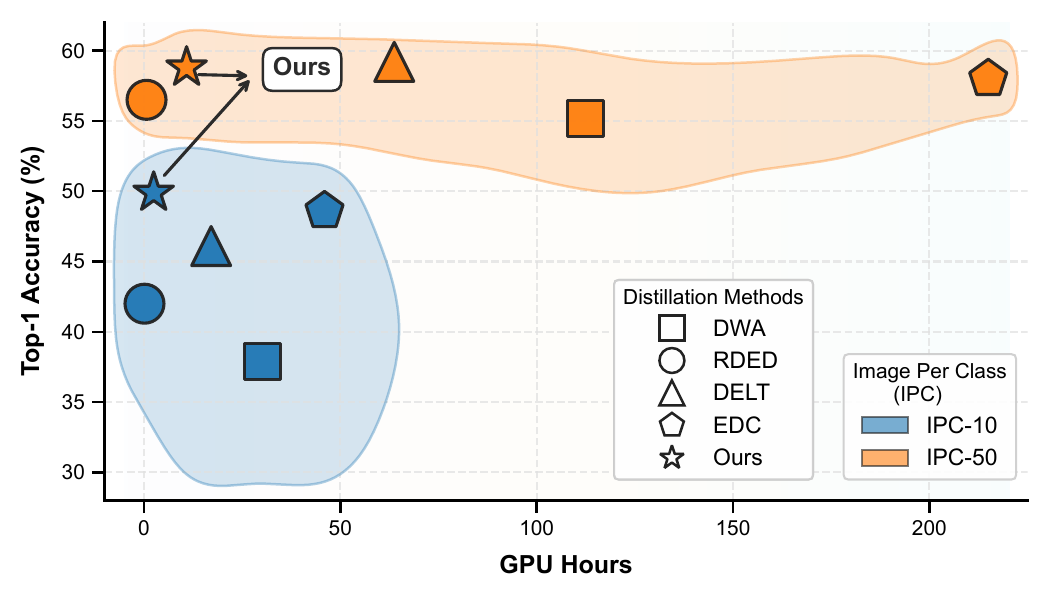}
    \centering
\caption{
Comparison of Top-1 accuracy and synthesis time on ImageNet-1K using ResNet-18 for various dataset distillation methods at IPC 10 and IPC 50.
Synthesis time is measured on a single RTX A6000 GPU. 
Our method converges substantially faster and achieves the highest accuracy, leading to the best accuracy–efficiency trade‑off.
}
\label{fig:intro}
   
\end{figure}

In response, a line of work on efficiency and scalability introduced the \textit{decoupled paradigm}~\cite{bohdal2020flexible,yin2023squeeze}, which separates model training from synthetic data optimization to avoid costly bi-level updates. This paradigm emphasizes the generation of soft labels to provide richer supervision, transferring much of the computational burden from optimization to label generation, as one study emphasis “a label is worth a thousand images”~\cite{ NEURIPS2024_Tian_label,xiao2025rethinkdc}. Building on this trend, RDED~\cite{sun2024diversity} advances the decoupled paradigm toward an optimization-free approach, eliminating iterative updates by directly extracting and recombining representative patches from real images, reducing synthesis time from hours to minutes.

Nevertheless, the drive toward faster distillation has not overcome the accuracy–efficiency trade‑off. Optimization-free methods are fast but lose accuracy, since the teacher only guides patch selection without optimizing the synthetic data. In contrast, optimization‑based methods typically achieve higher accuracy, especially when combined with optimization‑free strategies as an initialization step~\cite{Shao_NEURIPS2024_EDC,NRR-DD-2025}, but remain computationally expensive. For instance, EDC~\cite{Shao_NEURIPS2024_EDC}, a recent dataset distillation method, requires over \textbf{200 GPU hours} to distill ImageNet-1K at 50 IPC. Our work aims to bridge this gap by asking two key questions.

\begin{itemize}
\item \textbf{Q1:} How can we accelerate decoupling-based distillation to narrow the accuracy–efficiency gap?
\item \textbf{Q2:} Building on the efficiency focus, we ask whether decoupling-based distillation can reach its best accuracy early, such that further optimization not only adds cost but also degrade performance, challenging the assumption that more optimization always helps?
\end{itemize}

We attribute inefficiency in recent decoupling-based dataset distillation methods to redundancy which we argue stems from two main sources: similar patches generated during initialization, and repeated low-value updates during optimization. Existing methods propagate redundancy by applying uniform gradient updates across all regions, regardless of their contribution to loss reduction. This uniform treatment~\cite{yin2023squeeze,dwa2024neurips,Shao_NEURIPS2024_EDC} inflates computation without improving representational quality. Unlike prior work that focuses only on diversity~\cite{dwa2024neurips, Shao_2024_CVPR_Generalized, shen2024deltsimplediversitydrivenearlylate}, we explicitly \emph{reduce redundancy}, aligning efficiency and diversity to steer dataset distillation away from brute-force optimization toward leaner, more focused updates.

\begin{figure}[t]
    \centering


    \vspace{3pt}

    \begin{minipage}{\linewidth}
        \centering
        \includegraphics[width=\linewidth]{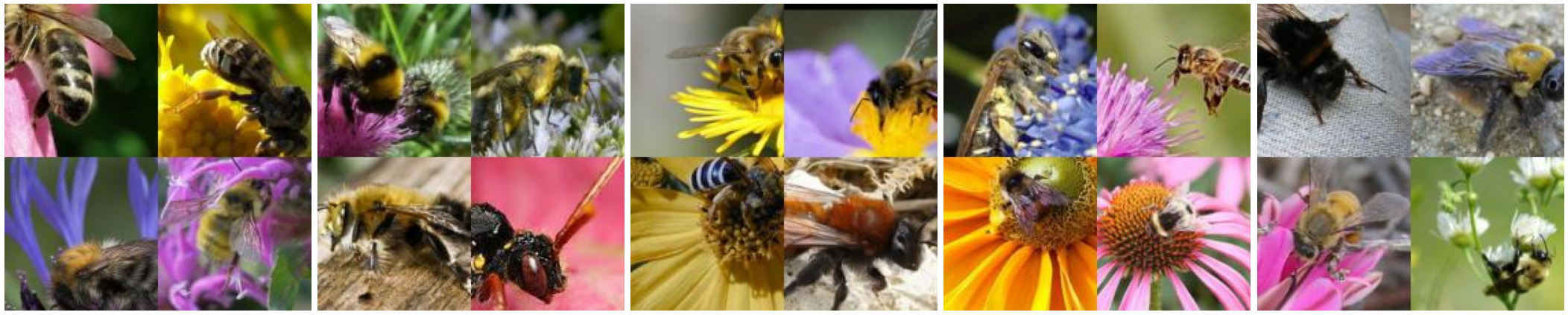}\\
        \vspace{1pt}
        \small (a) RDED
    \end{minipage}

    \vspace{3pt}

     \begin{minipage}{\linewidth}
        \centering
        \includegraphics[width=\linewidth]{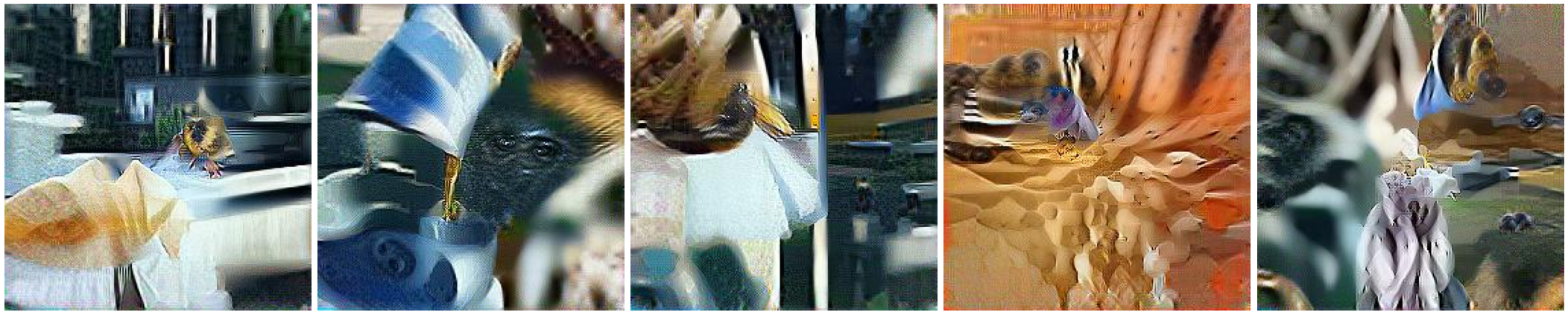}\\
        \vspace{1pt}
        \small (b) DELT
    \end{minipage}

    \vspace{3pt}
    
    \begin{minipage}{\linewidth}
        \centering
        \includegraphics[width=\linewidth]{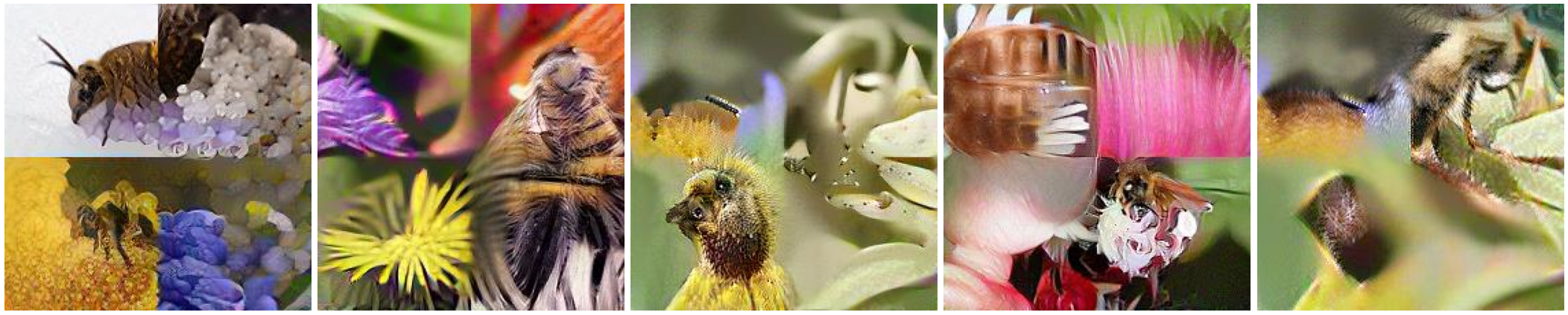}\\
        \vspace{1pt}
        \small (c) EDC
    \end{minipage}

    \vspace{3pt}

    \begin{minipage}{\linewidth}
        \centering
        \includegraphics[width=\linewidth]{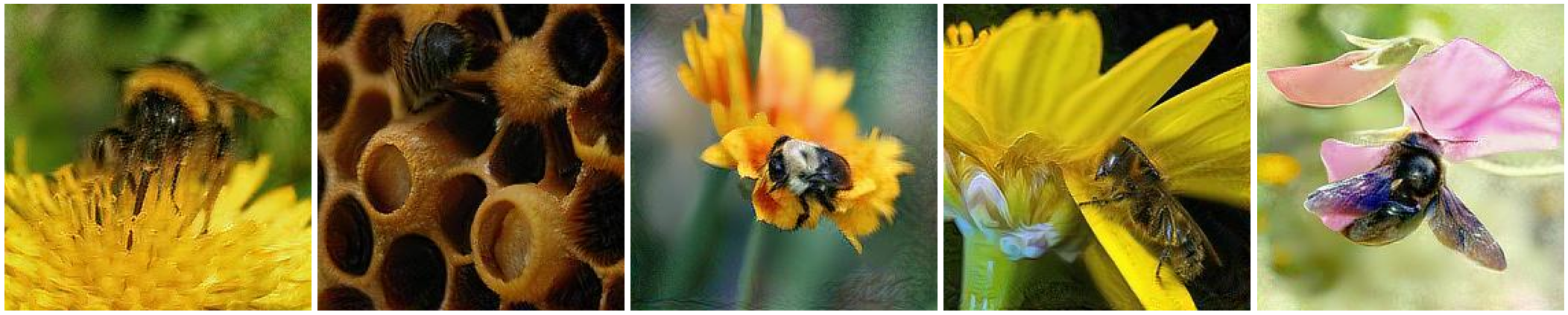}\\
        \vspace{1pt}
        \small (d) Ours
    \end{minipage}

    \vspace{4pt}

    \caption{Visual comparison of synthetic data generated by SRe$^2$L, RDED, DELLT, EDC, and our method E$^2$D, which produces more diverse, less redundant samples and preserves semantic integrity with full‑size feature representations.
}
    \label{fig:Visual_comparison}
\end{figure}

To address \textbf{Q1}, we revisit patch‑based initialization adopted in prior works~\cite{sun2024diversity,Shao_NEURIPS2024_EDC,shen2024deltsimplediversitydrivenearlylate}. This mechanism often produces clusters of similar patches, which reduces diversity and limits the representational coverage of the synthetic dataset, particularly when multiple patches originate from the same image, as shown in~\cref{fig:Visual_comparison}. Moreover, compact patch representations can distort features, compromising the quality of the synthetic data~\cite{zhong2025efficientdatasetdistillationdiffusiondriven}. We instead use full-size image initialization, which ensures better initial representations, reducing the need for extensive optimization. Remarkably, this simple change produces distilled data that already matches the accuracy of state-of-the-art methods before any optimization, substantially narrowing the efficiency–accuracy gap.

Next, we propose a two‑phase optimization strategy that reduces the redundant uniform updates of prior works. Rather than treating all regions in the synthetic data as equally valuable, the \textbf{exploration} phase broadly updates diverse regions to ensure coverage, before the \textbf{exploitation} phase concentrates updates on high-loss regions that provide stronger learning signals. This accelerates convergence and, with an accelerated student training schedule, delivers up to $20\times$ efficiency gains (\cref{fig:intro}).

\begin{figure}[t]
    \centering
    \includegraphics[width=0.95\linewidth]{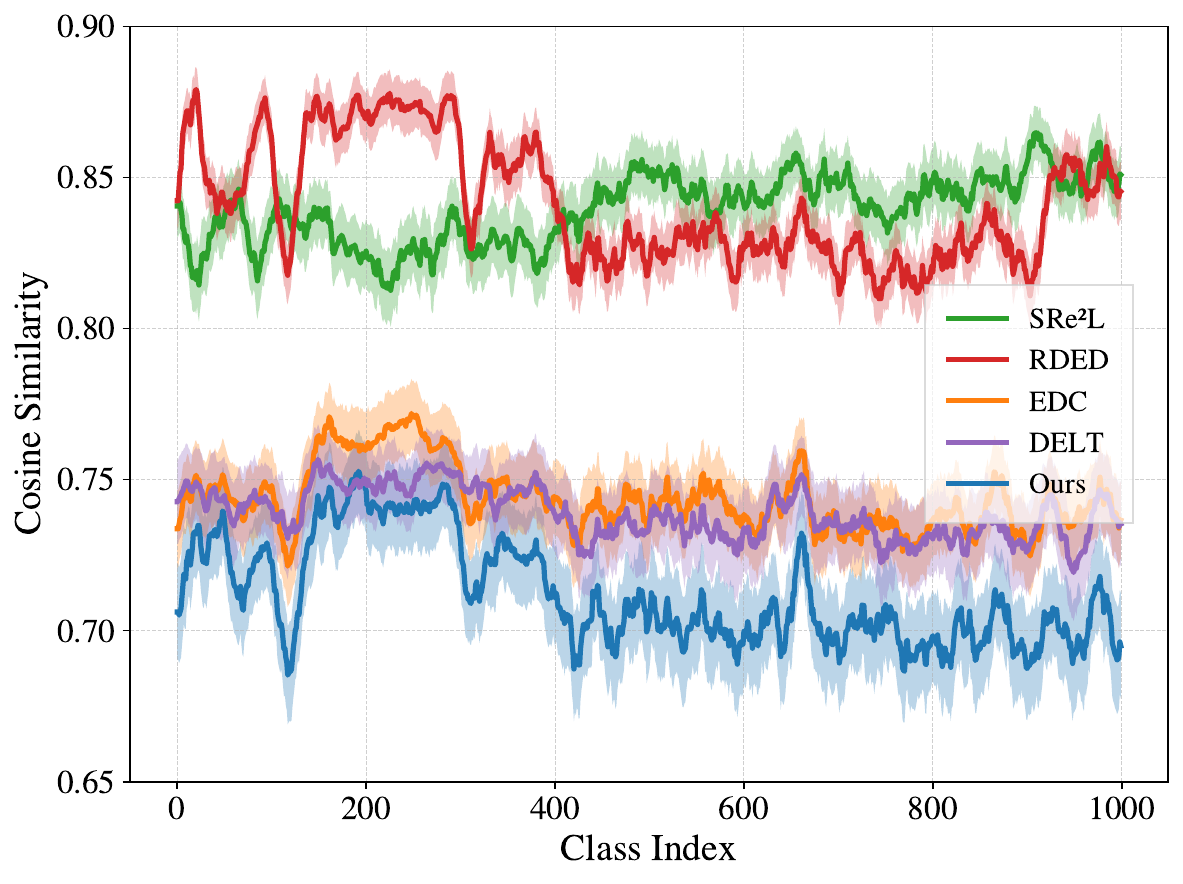}
    \caption{Semantic cosine similarity across ImageNet‑1K classes at IPC 50 using a ResNet‑18 teacher. Lower values indicate greater diversity and reduced redundancy; our method consistently achieves the lowest similarity.}   
    \label{fig:cosine_comparison}
\end{figure}

Turning to \textbf{Q2}, we make the counter-traditional observation that more optimization is not always beneficial. Prolonged updates tend to reinforce redundant global dataset statistics while eroding both diversity and fine-grained features inherent in the original dataset. In contrast, our approach demonstrates that a focused optimization strategy can yield higher-quality distilled datasets with fewer optimization steps. As illustrated in~\cref{fig:cosine_comparison}, our method achieves consistently lower semantic cosine similarity across ImageNet‑1K classes, indicating richer diversity. At the same time, it achieves state‑of‑the‑art accuracy, reaching peak ImageNet‑1K performance with $\approx10\times$ fewer optimization steps than EDC (\cref{fig:intro}). This challenges the conventional assumption that longer optimization is always beneficial and highlights the importance of efficient optimization.

Our main contributions are: 

\begin{enumerate}
    \item 
    We identify redundancy as a key inefficiency in recent decoupling-based dataset distillation. It arises from patch-based initialization, where similar crops dominate, and uniform optimization that ignores regional importance. While prior work assumes more optimization improves data, we find excessive updates amplify redundancy and degrade quality, reframing dataset distillation as an efficiency‑driven, diversity-aligned process.
    
    \item We propose the Exploration--Exploitation Distillation method (E$^2$D) which departs from prior uniform optimization methods by integrating full‑image initialization to preserve semantic integrity and diversity with a novel two‑phase optimization strategy that identifies high-loss regions and concentrates updates there to reduce redundancy and accelerate convergence.
    
    \item Extensive experiments validate the effectiveness of E$^2$D on large-scale benchmarks. On ImageNet-1K, it surpasses state-of-the-art methods while reducing synthesis time by up to \textbf{18×}. On ImageNet-21K, it yields accuracy gains of up to +9.6\% while remaining \textbf{4.3×} faster. These results show that E$^2$D achieves a superior balance between accuracy and computational cost, making it a practical solution for large-scale dataset distillation.

\end{enumerate}

\section{Related Work}
\label{Work}

\noindent\textbf{Bi-Level Optimization Distillation.}  
Early dataset distillation methods often adopt \textit{bi‑level optimization}, where synthetic data updates are aligned with updates of models trained on the original dataset, in a batch-by-batch manner. Matching strategies include \textit{gradient matching}~\cite{zhao2021DC,kimICML22_IDC,Acc-DD}, \textit{trajectory matching}~\cite{cazenavette2022distillation,cui2023scaling}, and \textit{distribution matching}~\cite{zhao2023DM,zhao2021DSA,Wang_2022_CVPR,zhou2022dataset,DSDM}. While distribution matching avoids costly bi‑level optimization, it typically underperforms gradient or trajectory matching. To address this trade-off, Zhang et al.~\cite{Acc-DD} accelerate gradient‑based optimization through early-model augmentation and parameter perturbation. DREAM~\cite{liu2023dream} accelerates optimization by matching only representative images from the original dataset. TESLA~\cite{cui2023scaling} extends trajectory matching to ImageNet‑1K with memory reduction and soft labels, but remains resource‑intensive. Overall, these methods achieve strong alignment but struggle to scale beyond small datasets.

\noindent\textbf{Decoupled-based Distillation.}  
To improve scalability, decoupled-based approaches separate model training from synthetic data optimization. SRe$^2$L~\cite{yin2023squeeze} introduces a three‑stage pipeline: \textit{Squeeze} (pretrain with Batch Normalization (BN) statistics), \textit{Recover} (independent optimization of synthetic data with global (BN) alignment), and \textit{Relabel} (assign soft labels via a teacher model). CDA~\cite{yin2023dataset_CDA} scales to ImageNet‑21K and further enhances performance with curriculum‑based augmentation. RDED~\cite{sun2024diversity} follows a similar decoupled approach but removes the optimization stage, yielding several-fold efficiency improvement by selecting teacher-guided patches that balance diversity and realism~\cite{li2022awesome}.

\noindent\textbf{Diversity in Decoupled-based Distillation.}
Several works address the limited diversity of global-statistics alignment in the decoupled-based distillation. DWA~\cite{dwa2024neurips} promotes diversity by dynamically adjusting teacher weights and separating BN mean and variance terms to strengthen variance alignment. G-VBSM~\cite{Shao_2024_CVPR_Generalized} expands diversity by incorporating multiple backbones, layers, and statistics. More recently, EDC~\cite{Shao_NEURIPS2024_EDC} builds on this framework to enhance both performance and efficiency through pipeline refinements across the distillation process. DELT~\cite{shen2024deltsimplediversitydrivenearlylate} enhances diversity at the IPC level by assigning different optimization depths to each synthetic image, producing image‑level variation to reduce uniformity in decoupled distillation

\noindent\textbf{Generative Distillation.}  
A parallel line of work explores generative approaches such as diffusion-based distillation~\cite{gu2024Minimax,Su_2024_CVPR_D4M,Qi_2025_CVPR,zhong2025efficientdatasetdistillationdiffusiondriven} and GAN-guided synthesis~\cite{zhao2022gan,cazenavette2023glad,zhong2024hierarchical} which shift from matching updates to training generative models that output synthetic data. However, they are orthogonal to optimization-based methods, as their generative outputs can be used as initialization for further refinement.

\section{Methodology}
\label{sec:Methodology}

\begin{figure*}[t]
    \centering
    \includegraphics[width=\linewidth]{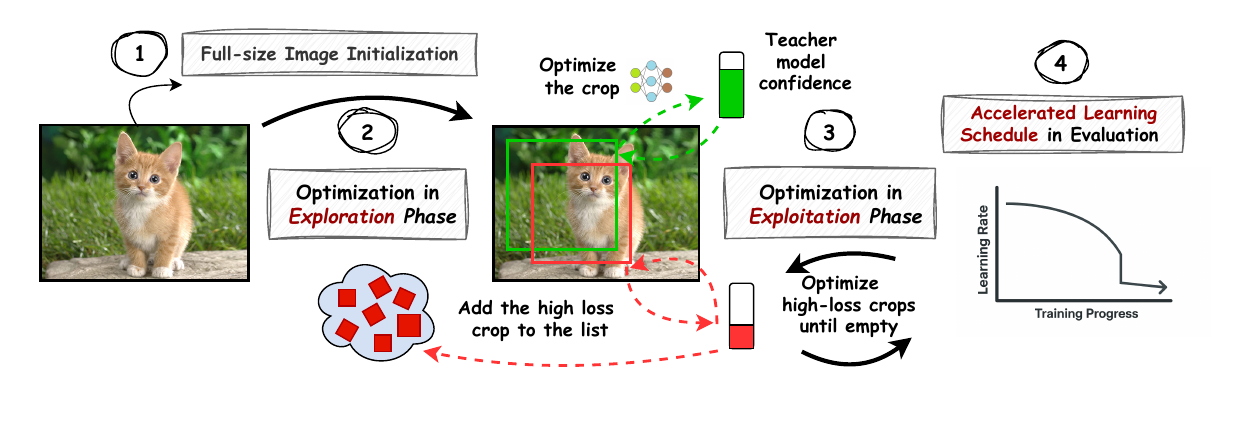}
    \vspace{-8mm}
    \caption{
        Overview of our proposed method E$^2$D. 
        The pipeline consists of four components: 
        (1) \textbf{Full-size Image Initialization}, which preserves the semantic and structural information of the original data, preventing distortion or redundancy; 
        (2) \textbf{Exploration Phase}, which identifies challenging high-loss regions and ensure balanced optimization;
        (3) \textbf{Exploitation Phase}, which iteratively refines these challenging regions for efficient convergence; and 
        (4) \textbf{Accelerated Learning Schedule}, applied during student training to further speed up convergence.
        Together, these components enable fast and effective dataset distillation with minimal redundancy.
    }
    \label{fig:main_method}
\end{figure*}

\subsection{Preliminaries}

Let $\mathcal{D} = \{(x_i, y_i)\}_{i=1}^N$ be the original training dataset, where $x_i \in \mathbb{R}^{H \times W \times C}$ is an image, and $y_i \in \{1, \ldots, L\}$ is its ground-truth label. Dataset distillation aims to synthesize a compact dataset $\mathcal{S} = \{(\tilde{x}_j, \tilde{y}_j)\}_{j=1}^M$ with $M \ll N$, such that a model trained on $\mathcal{S}$ achieves competitive performance compared to training on $\mathcal{D}$, by effectively capturing essential and diverse information within $\mathcal{D}$.

\begin{figure}[!ht]
    \centering
    \includegraphics[width=\linewidth]{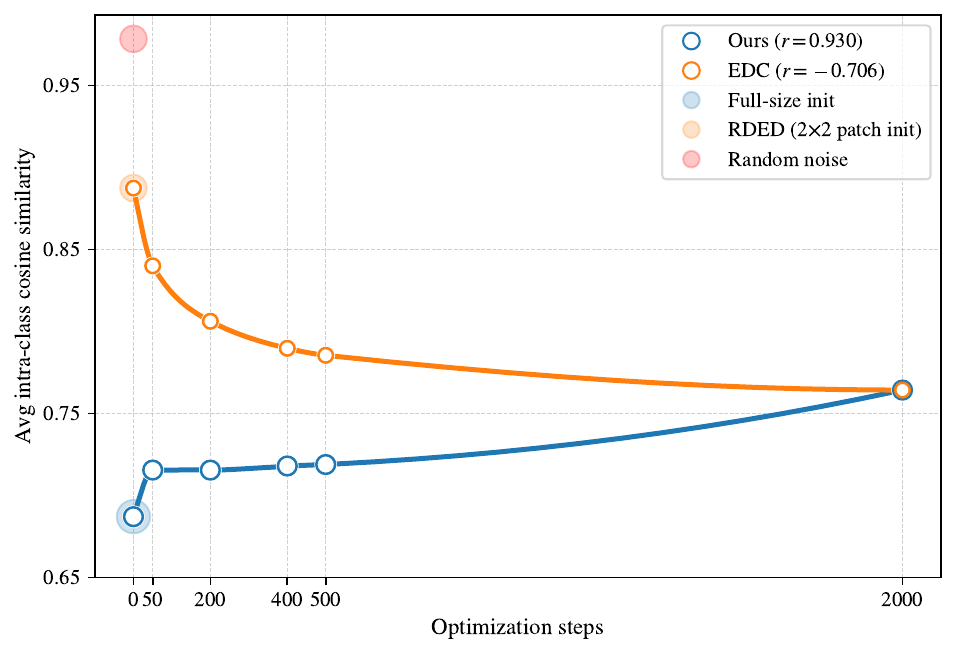}
    \caption{Cosine similarity trends across optimization steps.}
    \label{fig:avg_cosine_comparison}
\end{figure}

\subsection{Overview}

We study efficiency bottlenecks in recent decoupled dataset distillation methods and identify redundancy in two aspects of the pipeline: (i) suboptimal initialization that burdens the optimization with unnecessary corrective updates, and (ii) prolonged optimization that increasingly reinforces redundant signals and constrains efficiency. While prior work has primarily focused on improving diversity through architectural choices or data sampling strategies, we observe that excessive optimization signals not only increase synthesis cost but can also degrade instance-level feature richness, leading to negative returns. Our analysis shows that full‑image initialization provides a substantially stronger starting point than the patch‑based schemes commonly used in prior work~\cite{sun2024diversity, Shao_NEURIPS2024_EDC,shen2024deltsimplediversitydrivenearlylate,cui2025datasetdistillationcommitteevoting,NRR-DD-2025}. Building on these insights, we propose a novel optimization strategy that mitigates redundancy, preserves diversity and accelerate convergence. 

\subsection{Redundancy in Initialization and Optimization}
We treat redundancy as the degree of similarity between synthetic instances, viewing it as the opposite of diversity. Since diversity is inherently multi-dimensional, we adopt a practical proxy based on pairwise cosine similarity of teacher features. Higher similarity indicates more redundant representations, whereas lower similarity reflects greater instance-level diversity. Using this measure, we examine how teacher feature representations evolve under patch-based and full-image initialization, and how subsequent optimization amplifies or reduces redundant signals.

As shown in \cref{fig:avg_cosine_comparison}, patch-based initialization exhibits high redundancy and distorted feature representations, behaving closer to random noise than to the original data distribution captured by full-image initialization. In this regime, global-statistics optimization acts as a corrective step, reducing redundancy by steering noisy and misaligned features toward meaningful class-level structure. In contrast, when optimization is applied to full-image initialization, where features already align with the data distribution, the same global objective behaves differently. Rather than correcting noise, continued global alignment progressively amplifies shared class-level signals, increasing redundancy and eroding instance-level diversity over time. This reveals a conditional behavior of global-statistics optimization: it is beneficial when representations are far from the target distribution, but increasingly redundant once coarse alignment has been achieved.

Motivated by these observations, we aim to accelerate synthesis by controlling redundancy during optimization while retaining strong class-level supervision. Rather than adding new objectives, we control when global updates are applied by allowing broad corrective optimization early and then limiting redundant updates once representations are aligned. This preserves instance-level diversity inherited from full-image initialization while improving synthesis efficiency.

\begin{algorithm}[t]
\caption{Exploration--Exploitation Distillation (E$^2$D)}
\label{alg:mem-crop}
\kwInput{Original dataset $\mathcal{D}$ , teacher model $\mathcal{T}$, loss threshold $\epsilon$, exploration iterations $K$, total iterations $T$}
\kwOutput{Updated synthetic dataset $\mathcal{S}$}

\ForEach{($x_i,y_i$) $ \sim \mathcal{S}$}{
\tcc{Exploration Phase}
    Initialize $\mathcal{M}_i \leftarrow \varnothing$ \\
    Initialize $x_i$ with a random image from $\mathcal{D}$ \\
\For{$t = 1$ to $K$}{
        $\tilde{x}_i \gets \textit{RandomResizedCrop}(x_i)$\
        
        $l_i = \mathcal{L}_{ce}(\mathcal{T}(\tilde{x}_i), y_i)$\;
        
        \If{$l_i > \epsilon$}{
            Store $(\text{crop coordinates}, l_i)$ in $\mathcal{M}_i$\;
        }
            Update $\tilde{x}_i$ using the baseline distillation loss.}
        
\tcc{Exploitation Phase}

\For{$t = K+1$ to $T$}{
        \If{$\mathcal{M}_i \neq \emptyset$}{
            Sample $(\text{crop coordinates})$ from $\mathcal{M}_i$ via softmax over $l_i$\;
            
            $\tilde{x}_i \gets \text{Crop}(x_i, \text{crop coordinates})$\;
            
            $l_i' = \mathcal{L}_{ce}(\mathcal{T}(\tilde{x}_i), y_i)$\;
            
            \If{$l_i' > \epsilon$}{
                Update stored loss in $\mathcal{M}_i$\;
            }
            \Else{
                Remove crop from $\mathcal{M}_i$\;
            }
        }
        \Else{
            \tcc{Early Stopping}
            \textbf{break}
        }
       Update $\tilde{x}_i$ using the baseline distillation loss.

    }
}
\end{algorithm}

\subsection{Exploration--Exploitation Optimization Strategy}
We propose a two-phase optimization strategy that mitigates the redundancy of random multi-crop updates used in prior methods~\cite{yin2023squeeze,Shao_NEURIPS2024_EDC,dwa2024neurips}. In conventional multi-crop optimization, each iteration performs in-place updates on randomly sampled local regions, repeatedly visiting similar patches and generating redundant learning signals. Inspired by the \textit{exploration–exploitation} trade-off in reinforcement learning~\cite{kaelbling1996reinforcement}, our method separates this process into distinct exploration–exploitation phases to achieve efficient coverage and targeted refinement, directing computation toward regions with strong learning potential to increase information density and accelerate convergence. Our method is formally outlined in \cref{alg:mem-crop}.

\paragraph{Exploration Phase.}
This phase performs random multi-crop optimization over $K$ iterations to broadly update the synthetic data and record informative regions. 
For each synthetic image $x_i$, crops that yield high teacher loss $L_{\text{ce}}(T(\tilde{x}_i), y_i) > \epsilon$ are stored with their coordinates and loss values in a per-image memory buffer $M_i$. 
These buffers accumulate the teacher’s uncertainty landscape, identifying regions that are under-optimized or poorly aligned with class semantics. 
The same random crop is applied to all images in the batch to maintain computational efficiency.

\paragraph{Exploitation Phase.}
After $K$ exploration iterations, optimization shifts to focused refinement of the high-loss regions identified in the exploration phase.
For each synthetic image \( x_i \),  crops are sampled from its buffer $M_i$ with probabilities proportional to their stored losses:

\begin{equation}
p_{ij} = \frac{\exp(\ell_{ij})}{\sum_{k=1}^{N_i} \exp(\ell_{ik})}
\label{eq:softmax}
\end{equation}

where \( p_{ij} \) denotes the probability of selecting the \( j \)-th high-loss crop in image \( x_i \), 
\( \ell_{ij} \) represents the loss of the \( j \)-th crop in image \( x_i \), and \( N_i \) indicates the total number of crops stored in the buffer \( \mathcal{M}_i \).

This softmax weighting prioritizes difficult regions while maintaining diversity among selected crops. 
Updated losses are re-evaluated and refreshed in memory; regions that fall below the loss threshold are discarded, preventing redundant computation on well-optimized areas.

Optimization proceeds until all buffers are emptied or the maximum iteration budget $T$ is reached. Early stopping prevents over-optimization of the distilled data, preserving image features inherited from the original dataset.

\begin{table*}[t]
\begin{adjustbox}{width=1\linewidth}
\begin{tabular}{@{}cc|cccccccc@{}}
\toprule
\multicolumn{2}{c|}{} & \multicolumn{8}{c}{ResNet-18} \\
\cmidrule(lr){3-10}
Dataset & IPC & SRe$^2$L & CDA & DWA & RDED & DELT & EDC & Ours$_{\text{(optimization-free)}}$ & Ours$_{\boldsymbol{10\times}}$ \\\midrule
ImageNet-1K & 1 & -- & -- & -- & 6.6 ± 0.2 & -- & 12.8 ± 0.1 & 11.5 ± 0.4 & \cellcolor{mygray} \textbf{12.9 ± 0.4} \\
& 10 & 21.3 ± 0.6 & 31.4 ± 0.5 & 33.5 ± 0.3 & 37.9 ± 0.2 & 46.1 ± 0.4 & 48.6 ± 0.3 & 48.1 ± 0.2 & \cellcolor{mygray} \textbf{50.0 ± 0.1} \\
& 50 & 46.8 ± 0.2 & 51.8 ± 0.4 & 53.5 ± 0.3 & 55.2 ± 0.2 & \cellcolor{mygray} \textbf{59.2 ± 0.4} & 58.0 ± 0.2 & 58.4 ± 0.1 & 58.9 ± 0.1 \\
\bottomrule
\end{tabular}
\end{adjustbox}
\caption{\textbf{Comparison of dataset distillation methods on ImageNet-1K using ResNet-18 as the evaluation model.} Our method is presented in two variants: with an acceleration factor of \textit{$\boldsymbol{10\times}$} and an optimization-free variant. Our results are reported as mean ± standard deviation over five evaluation models, each randomly initialized and trained independently on the distilled dataset. Missing entries (--) indicate results not reported in the original work.}
\label{tab:imagenet_1k_results}
\end{table*}

\begin{table*}
\centering
\begin{adjustbox}{width=0.77\linewidth}
\begin{tabular}{@{}cc|ccccc@{}}
\toprule
\multicolumn{2}{c|}{} & \multicolumn{5}{c}{ResNet-18} \\
\cmidrule(lr){3-7}
Dataset & IPC & SRe$^2$L & CDA & D3S & Ours$_{\text{(optimization-free)}}$ & Ours$_{\boldsymbol{5\times}}$ \\\midrule
ImageNet-21K & 10 & 18.5 ± 0.2 & 22.6 ± 0.2 & 26.9 ± 0.1 & 28.8 ± 0.02 & \cellcolor{mygray} \textbf{32.1 ± 0.4} \\
& 20 & 21.8 ± 0.1 & 26.4 ± 0.1 & 28.5 ± 0.1 & 33.9 ± 0.1 & \cellcolor{mygray} \textbf{36.0 ± 0.1} \\
\bottomrule
\end{tabular}
\end{adjustbox}
\vspace{0.5em}
\caption{\textbf{Comparison of dataset distillation methods on ImageNet-21K using ResNet-18 as the evaluation model.} Our method is presented with an acceleration factor of \textit{$\boldsymbol{5\times}$}, along with an optimization-free variant. Our results are reported as mean ± standard deviation over three evaluation models, each randomly initialized and trained independently on the distilled dataset.}
\label{tab:imagenet_21k_results}
\end{table*}

\section{Experiments}
\label{sec:Experiment}

To evaluate our method, we perform experiments on two large-scale datasets using multiple model architectures to assess its effectiveness and its ability to generalize across architectures. We conduct comprehensive ablations and convergence analyses, examining how varying acceleration factors, including optimization-free settings affect the efficiency–performance trade-off.

\subsection{Experimental Setups}
\noindent\textbf{Datasets.}
We evaluate our method on two large-scale and computationally intensive benchmarks: ImageNet-1K ~\cite{imagenet} and ImageNet-21K-P (referred to as ImageNet-21K throughout the paper)~\cite{ImageNet-21K}. ImageNet-1K consists of 1,000 classes and 1.28 million images, while ImageNet-21K includes 10,450 classes and over 11 million images. We omit results on small-scale datasets (e.g., CIFAR-10/100), where our baseline EDC already achieves state-of-the-art performance with very few optimization iterations, which makes further acceleration provide negligible efficiency gains. Additional discussion is provided in the appendix.

\noindent\textbf{Network architectures.}
For \cref{tab:imagenet_1k_results} and \cref{tab:imagenet_21k_results}, we report results using the widely adopted ResNet-18 architecture as the evaluation model.
Following previous works~\cite{Shao_NEURIPS2024_EDC,yin2023dataset_CDA}, and to ensure the distilled data are not overfitted to any single architecture, we further assess cross-architecture generalization on diverse models including ResNet-\{50,101\}~\cite{resnet},  MobileNet-V2~\cite{sandler2019mobilenetv2invertedresidualslinear},  EfficientNet-B0~\cite{EfficientNet}, ConvNext-Tiny~\cite{liu2022convnet}, ShuffleNet-V2~\cite{Ma_2018_ECCV}, DeiT-Tiny~\cite{touvron2021trainingdataefficientimagetransformers}, RegNet-Y-8GF~\cite{Radosavovic_2020_CVPR}, and DenseNet-121~\cite{8099726}.

\noindent \textbf{Baselines.}
We compare against recent state-of-the-art methods in large-scale dataset distillation. For ImageNet-1K, we compare with SRe$^2$L~\cite{yin2023squeeze}, CDA~\cite{yin2023dataset_CDA}, DWA~\cite{dwa2024neurips}, G-VBSM~\cite{Shao_2024_CVPR_Generalized}, RDED~\cite{sun2024diversity}, DELT~\cite{shen2024deltsimplediversitydrivenearlylate}, and EDC~\cite{Shao_NEURIPS2024_EDC}. On ImageNet-21K, we consider SRe$^2$L, CDA, and D3S~\cite{D3S} as baselines.

\noindent \textbf{Implementation details.}
We adopt EDC as the implementation backbone for our method in ImageNet-1K and CDA in ImageNet-21K. Both are accelerated using our Exploration–Exploitation strategy, allocating 70\% of iterations to exploration and the remaining 30\% to exploitation.  Our method integrates seamlessly with state-of-the-art methods, demonstrating its adaptability cross both large-scale datasets and different distillation methods. EDC and CDA are optimized for 2000 iterations, and our reported acceleration factors are measured relative to this budget. When extending our method to ImageNet-21K, we reused the optimization hyperparameters from prior work. Under this setup, even slight optimization caused performance to degrade below that of the optimization-free variant. We attribute this to a distributional mismatch between the teacher model and the synthesized data. 
To mitigate this, we reduced the learning rate and relaxed the batch-normalization regularization term $\alpha_{\text{BN}}$, which stabilized optimization and yielded smoother convergence curves. Additional implementation details and hyperparameter configurations are provided in the appendix.

\begin{figure}[t]
    \centering
    \begin{subfigure}{\linewidth}
        \centering
        \includegraphics[width=\linewidth]{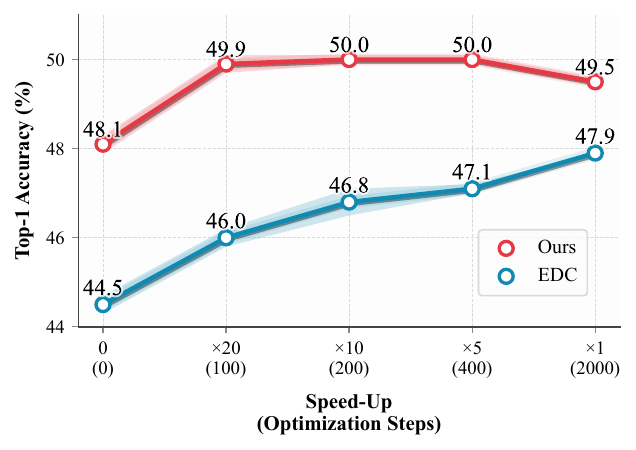}
        \caption{ImageNet-1K (IPC 10).}
    \end{subfigure}
    \vspace{0.8em}
    \begin{subfigure}{\linewidth}
        \centering
        \includegraphics[width=\linewidth]{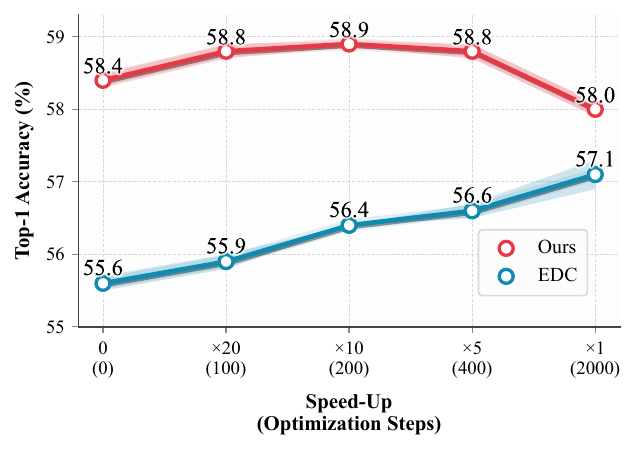}
        \caption{ImageNet-1K (IPC 50).}
    \end{subfigure}

    \caption{
        Top‑1 accuracy vs. speedup for our method and baseline EDC under two IPC settings. Our method converges faster and starts from a stronger no‑optimization starting point. In our method, extra optimization past convergence degrades performance, counter to the common assumption that more optimization is always better.
    }
    \label{fig:acc_vs_speed_all}
\end{figure}
\subsection{Results}

Our method demonstrates strong performance as shown in \cref{tab:imagenet_1k_results} and \cref{tab:imagenet_21k_results}. Despite being several folds more efficient, it consistently achieves superior accuracy compared to recent state-of-the-art methods.

On ImageNet-1K, our method sets a new benchmark at the highly compressed IPC = 10 setting, reaching 50\% accuracy. On the more challenging ImageNet-21K, it delivers a substantial improvement by achieving 32.1\% accuracy at IPC = 10. While the performance gap narrows as IPC increases, our method continues to outperform state-of-the-art methods, achieving 58.9\% on ImageNet-1K at IPC = 50 and 36\% on ImageNet-21K at IPC = 20.
Our optimization-free variant performs on par with the strongest prior method, EDC, on ImageNet-1K, despite requiring no optimization. Remarkably, on ImageNet-21K, our optimization-free variant  outperforms all existing methods while using only half the distillation budget (IPC = 10) compared to state‑of‑the‑art methods trained at IPC = 20.

\begin{table*}[!ht]
  \centering
  \begin{adjustbox}{width=1.0\linewidth}
  \begin{tabular}{cc|ccccccc}
    \toprule
    \multirow{2}{*}{IPC} & \multirow{2}{*}{Method} & \multicolumn{7}{c}{Evaluation model} \\
    {} & {} & ResNet-18 & ResNet-50 & ResNet-101 & MobileNet-V2 & EfficientNet-B0 & ConvNext-Tiny & ShuffleNet-V2 \\
    \midrule
     & RDED               & 42.0 & 46.0 & 48.3 & 34.4 & 42.8 & 48.3 & 19.4 \\
    10 & EDC                & 48.6 & 54.1 & 51.7 & 45.0 & 51.1 & 54.4 & 29.8 \\
    \rowcolor{mygray}
     & $\text{Ours}_{10\times}$ & \textbf{50.0} & \textbf{55.6} & \textbf{54.7} & \textbf{46.9} & \textbf{52.9} & \textbf{55.0} & \textbf{45.4} \\
    \midrule
     & RDED               & 56.6 & 63.7 & 61.2 & 53.9 & 57.6 & 65.4 & 30.9 \\
    50 & EDC                & 58.0 & 64.3 & 64.9 & 57.8 & 60.9 & 66.6 & 45.7 \\
    \rowcolor{mygray}
     & $\text{Ours}_{10\times}$ & \textbf{58.9} & \textbf{64.7} & \textbf{65.1} & \textbf{58.1} & \textbf{61.3} & \textbf{66.8} & \textbf{56.4} \\
    \bottomrule
  \end{tabular}
  \end{adjustbox}
  \vspace{0.5em}
  \caption{Cross-architecture generalization performance on ImageNet-1K at IPC 10 and 50 across multiple evaluation models.}
  \label{tab:crossarch_1K}
\end{table*}

\begin{table*}[!ht]
  \centering
  \begin{adjustbox}{width=0.9\linewidth}
  \begin{tabular}{cc|cccccc}
    \toprule
    \multirow{2}{*}{IPC} & \multirow{2}{*}{Method} & \multicolumn{6}{c}{Evaluation model} \\
    {} & {} & ResNet-18 & ResNet-50 & ResNet-101 & DenseNet-121 & RegNet-Y-8GF & ConvNeXt-Tiny \\
    \midrule
    \multirow{4}{*}{20} 
      & SRe$^2$L             & 21.8 & 31.2 & 33.2 & 24.6 & 34.2 & 34.9 \\
      & CDA               & 26.4 & 35.3 & 36.1 & 28.6 & 36.1 & 36.3 \\
      & D3S               & 28.5 & 35.4 & 36.0 & 31.9 & 36.4 & --   \\
    \rowcolor{mygray}
      & $\text{Ours}_{5\times}$ & \textbf{36.0} & \textbf{36.9} & \textbf{37.3} & \textbf{33.1} & \textbf{37.4} & \textbf{37.2} \\
    \bottomrule
  \end{tabular}
  \end{adjustbox}
  \vspace{0.5em}
  \caption{Cross-architecture generalization performance on ImageNet-21K data at IPC 20 across multiple evaluation models. Missing entries (--) indicate unreported results with no publicly available codebase.}
  \vspace{-2mm}
  \label{tab:crossarch_21K}
\end{table*}

\subsection{Cross-architecture Generalization}
A key property of distilled datasets is their ability to generalize across diverse model architectures. We evaluate our method against its closest baselines on a range of architectures, as shown in \cref{tab:crossarch_1K} for ImageNet-1K and \cref{tab:crossarch_21K} for ImageNet-21K. Our method consistently outperforms state-of-the-art methods across all models, demonstrating improved robustness and consistently superior cross-architecture performance.

\subsection{Acceleration and Efficiency}

\noindent \textbf{Acceleration of Optimization.}
We study how accelerating our method influences performance relative to the EDC baseline on ImageNet-1K. As shown in \cref{fig:acc_vs_speed_all}, optimization-free settings generally perform increasingly close to the optimization-based ones as the IPC increases, a trend also observed in prior work~\cite{xiao2025rethinkdc}. Notably, our method in the optimization-free setting nearly matches the performance of the optimized version at high IPC, highlighting its efficiency given the substantial cost of optimization at large IPC. The 20$\times$ accelerated variant of our method results in less than 0.1\% accuracy drop, striking an effective balance between speed and performance. The 10$\times$ accelerated version of our method is sufficient to reach convergence, while the 5$\times$ version does not provide additional gains. As the number of optimization steps approaches the standard 2000 iterations, both our method and the baseline converge in performance, suggesting that excessive optimization reinforces redundant signals and reduces instance‑level diversity, supporting a counter‑traditional view that more optimization is not necessarily better.
\begin{table}[!ht]
\centering
\vspace{0.5em} 
\begin{subtable}[t]{\linewidth}
\centering
\begin{adjustbox}{width=1\linewidth}
\begin{tabular}{lccc}
\toprule
{} & Ours$_{20\times}$ & Ours$_{10\times}$ & EDC \\
\midrule
Synthesis Time (hrs) & 12.3  & 24.0 & 229.8  \\
Synthesis Speedup & $18.6\times$ & $9.5\times$ & $1.0\times$ \\
\bottomrule
\end{tabular}
\end{adjustbox}
\caption{ImageNet-1K (IPC = 50)}
\end{subtable}

\vspace{1em} 

\begin{subtable}[t]{\linewidth}
\centering
\begin{adjustbox}{width=1\linewidth}
\begin{tabular}{lccc}
\toprule
& Ours$_{10\times}$ & Ours$_{5\times}$ & CDA \\
\midrule
Synthesis Time (hrs) & 38.8 &  67.7 & 296.5 \\
Synthesis Speedup & $7.6\times$ & $4.3\times$ & $1.0\times$ \\
\bottomrule
\end{tabular}
\end{adjustbox}
\caption{ImageNet-21K (IPC = 20)}
\end{subtable}
\vspace{0.5em}
\caption{
\textbf{Synthesis time and acceleration factor comparison.} Each subtable reports synthesis time (hours) and speedup over baseline on a single NVIDIA RTX A6000 GPU with ResNet‑18, evaluated on different datasets.
}
\label{tab:synthesis_time}
\end{table}

\noindent \textbf{Efficiency Comparison.}
We compute the actual acceleration factor of our method, accounting for the overhead introduced by the Exploration–Exploitation strategy. To ensure fair comparison, all experiments are conducted on a single NVIDIA RTX A6000 GPU, and synthesis time is reported excluding data and model loading overhead.

As shown in \cref{tab:synthesis_time}, our method accelerates EDC by over \textbf{18$\times$} while maintaining higher near-converged accuracy. The additional cost of the Exploration–Exploitation optimization strategy remains modest relative to the \textbf{2$\times$} improvement in synthesis efficiency (see \cref{tab:exp_exploit} in the study). Beyond synthesis speed, we also address memory efficiency by applying Automatic Mixed Precision (AMP) to the forward pass of the bottleneck backbone architecture, EfficientNet-B0. This tweak reduces peak GPU memory consumption from 16.05 GB to 14 GB without any measurable loss in accuracy, allowing our method to fit comfortably on a single 16 GB GPU. 

For ImageNet-21K, our method also improves CDA efficiency, though the gain is smaller due to the dataset’s large-scale, which increases the cost of maintaining high-loss crop buffers compared to random cropping. Nevertheless, our method achieves consistently faster synthesis with stronger accuracy across both benchmarks.

\subsection{Discussion}
\label{subsec:Discussion}

\noindent \textbf{Effectiveness of Exploration–Exploitation Strategy}

We evaluate the impact of our Exploration–Exploitation strategy against the random multi-crop optimization used in the baselines. As shown in \cref{tab:exp_exploit}, our method outperforms the baseline even when the latter is given twice the optimization budget, enabling up to 2× faster convergence during synthesis.

\begin{table}[!ht]
  \centering
  \begin{adjustbox}{width=\linewidth}
  \begin{tabular}{c|ccc}
    \toprule
    IPC & Ours$_{20\times}$ (R) & Ours$_{20\times}$ (E$^2$) & Ours$_{10\times}$ (R) \\
    \midrule
    10 & 49.7 \std{0.2} & \textbf{49.9 \std{0.1}} & 49.8 \std{0.1} \\
    50 & 58.4 \std{0.1} & \textbf{58.8 \std{0.04}} & 58.8 \std{0.1} \\
    \bottomrule
  \end{tabular}
  \end{adjustbox}
  \vspace{0.5em}
  \caption{\textbf{Effectiveness of Exploration–Exploitation strategy.} Ablation of our Exploration–Exploitation optimization (E$^2$) vs. random multi-crop optimization (R) on ImageNet-1K. E$^2$ improves accuracy and enables up to 2$\times$ faster convergence during synthesis.}
  \label{tab:exp_exploit}
\end{table}

\noindent  \textbf{Design Space of E\textsuperscript{2}D.}

To examine the design space of our exploration–exploitation method, we evaluate four variants in \cref{tab:design_space}. \textbf{(1) Exploit only.} Image updates occur exclusively during exploitation, while exploration evaluates random crops using the teacher loss without performing any image-level updates. \textbf{(2) GradCAM-guided probing.} For each synthetic image, a GradCAM map is computed, and crops are sampled inversely to the activation magnitude so that low-activation regions are prioritized. \textbf{(3) Alternating cycles.} A periodic schedule of twenty steps in which exploration and exploitation alternate, replacing our two-stage procedure. \textbf{(4) Random multi-crop optimization.} The default strategy adopted in prior work, where crops are sampled uniformly at random throughout optimization. Across these variants, our method achieves the strongest performance. The drop in the exploit-only setting highlights the importance of uniform image-level updates, and the GradCAM variant further shows that semantic activation maps alone are insufficient, performing worse than simple random cropping.

\begin{table}[!ht]
  \centering
  \begin{adjustbox}{width=\linewidth}
  \begin{tabular}{c|ccccc}
    \toprule
    IPC & Exploit-Only & Grad-CAM & Alt. Cycles & Random Crop & E$^2$D \\
    \midrule
    10 & 48.5 \std{0.2} & 49.4 \std{0.2} & 49.6 \std{0.2} & 49.7 \std{0.2} & \textbf{49.9 \std{0.1}} \\
    \bottomrule
  \end{tabular}
  \end{adjustbox}
  \vspace{0.4em}
  \caption{Ablation of the design space of E$^2$D under a $\times$20 acceleration factor on ImageNet-1K (IPC 10).}
  \label{tab:design_space}
\end{table}

\noindent \textbf{When to Switch to Exploitation Phase} 

We analyze the effect of switching from exploration to exploitation at different points, controlled by hyperparameter $K$. If $K$ is too small, the switch is premature and crop diversity suffers; if too large, it degenerates to random cropping. Our experiments show that performance remains stable across a broad range of $K$ values, suggesting robustness to the exact choice of $K$. As shown in \cref{tab:ablation_k}, allocating 70\% of the iterations to exploration and 30\% to exploitation yields the best accuracy, offering an effective balance between crop diversity and targeted refinement.

\begin{table}[!ht]
\centering
\begin{adjustbox}{width=1.0\linewidth}
\begin{tabular}{c|cccc}
\toprule
\textbf{K} & 40 & 60 & 70 & 80 \\
\midrule
\textbf{Ours$_{20\times}$} 
  & 49.7 \std{0.2} 
  & 49.8 \std{0.1} 
  & \textbf{49.9 \std{0.1}} 
  & 49.7 \std{0.1} \\
\bottomrule
\end{tabular}
\end{adjustbox}
\vspace{0.7em}
\caption{\textbf{When to switch to exploitation}. Ablation study on the number of exploration iterations $K$ (out of 100 total iterations) for our method accelerated by $20\times$ on ImageNet-1K with IPC = 10.}
\label{tab:ablation_k}
\end{table}

\section{Conclusion}
We introduced Exploration--Exploitation Distillation \textbf{(E$^2$D)}, a redundancy-reducing method for efficient large-scale dataset distillation. E$^2$D integrates full-image initialization, targeted Exploration–Exploitation optimization, and an accelerated student schedule to minimize redundant updates while preserving diversity. Evaluated on two large‑scale benchmarks, ImageNet‑1K and ImageNet‑21K, E$^2$D achieves higher accuracy with substantially lower synthesis cost. These findings show that greater accuracy does not require more optimization; rather, directing computation where it matters most leads to better accuracy–efficiency trade-offs in large-scale dataset distillation.

\noindent
\textbf{Limitations and Future Work.} 
Our method does not account for the cost/efficiency of the relabeling stage or of constructing or selecting the pretrained models and collecting their statistics. We rely on simple random sampling for initialization, but more robust strategies may lead to more consistent performance. Furthermore, bridging the accuracy–efficiency gap at higher IPC levels may require refined optimization and initialization strategies beyond existing approaches.

\section*{Acknowledgments}
This work was partially supported by the National Science Foundation (NSF) grants BCS-2416846, OAC-2417850, and DUE-2526340. Muhammad J.~Alahmadi is supported in part by King Abdulaziz University (KAU). This research used resources of the Oak Ridge Leadership Computing Facility at Oak Ridge National Laboratory, supported by the Office of Science of the U.S.\ Department of Energy under Contract No.~DE-AC05-00OR22725. The views and conclusions are those of the authors and should not be interpreted as representing the official policies of the funding agencies or the government.

{
    \small
    \bibliographystyle{ieeenat_fullname}
    \bibliography{main}
}

\clearpage
\setcounter{page}{1}
\maketitlesupplementary

\vspace{0.5em}
\begin{table*}[!ht]
\centering
\begin{tabular}{ll@{\hspace{1cm}}ll}
\toprule
\multicolumn{2}{c}{\textbf{Synthesis}} & \multicolumn{2}{l}{\textbf{Relabeling \& Evaluation}} \\
\midrule
Iteration              & 200               & Epochs                & 300 \\
Optimizer              & \begin{tabular}[t]{@{}l@{}}AdamW\\$\beta_1 = 0.5$, $\beta_2 = 0.9$\end{tabular} 
                      & Optimizer         & AdamW \\
                      &                   & Loss Type             & MSE-GT + 0.025·CE \\
Learning rate          & 0.05              & Learning rate         & 0.001 \\
Batch Size             & 80                & Batch Size            & 100 \\
$K$, $\epsilon$        & 140, 0.5          & LR Schedule           & \begin{tabular}[t]{@{}l@{}}SSRS, $\zeta = 2$ \\SmoothingLR at IPC = 1\end{tabular} \\
Initialization         & \begin{tabular}[t]{@{}l@{}}Random sampling\\(RDED at IPC = 1)\end{tabular} 
                      & Augmentation      & \begin{tabular}[t]{@{}l@{}}RandomResizedCrop\\RandomHorizontalFlip\\CutMix\end{tabular} \\
$\alpha_{\text{BN}}$   & 0.1               & EMA Rate              & 0.99 \\
\bottomrule
\end{tabular}
\vspace{0.5em}
\caption{Hyperparameter settings for ImageNet-1K.}
\label{tab:hyperparams_imagenet1k}
\end{table*}

\vspace{0.8em}

\begin{table*}[!ht]
\centering
\begin{tabular}{ll@{\hspace{1cm}}ll}
\toprule
\multicolumn{2}{c}{\textbf{Synthesis}} & \multicolumn{2}{l}{\textbf{Relabeling \& Evaluation}} \\
\midrule
Iteration              & 400               & Epochs                & 300 \\
Optimizer              & \begin{tabular}[t]{@{}l@{}}AdamW\\$\beta_1 = 0.5$, $\beta_2 = 0.9$\end{tabular}
                      & Optimizer         & AdamW \\
                      &                   & Loss Type             & KL-Div \\
Learning rate          & 0.005             & Learning rate         & 0.005 \\
Batch Size             & 80                & Batch Size            & 100 \\
$K$, $\epsilon$        & 280, 0.3          & LR Schedule           & \begin{tabular}[t]{@{}l@{}}SSRS, $\zeta = 2$\end{tabular} \\
Initialization         & Random sampling   & Augmentation          & \begin{tabular}[t]{@{}l@{}}RandomResizedCrop\\RandomHorizontalFlip\\CutMix\end{tabular} \\
$\alpha_{\text{BN}}$   & 0.01              &                       & \\
\bottomrule
\end{tabular}
\vspace{0.5em}
\caption{Hyperparameter settings for ImageNet-21K.}
\label{tab:hyperparams_imagenet21k}
\end{table*}

\section{Further Discussion}

\noindent \textbf{Additional Cross‑architecture Evaluation.} 

\noindent To expand our evaluation of cross‑architecture generalization, \cref{tab:additional_crossarch_1K} reports results on transformer-based and hybrid models, where our method consistently achieves the strongest performance.

\begin{table}[!ht]
  \centering
  \begin{adjustbox}{width=1\linewidth}
  \begin{tabular}{cc|cccc}
    \toprule
    \multirow{2}{*}{IPC} & \multirow{2}{*}{Method} 
    & \multicolumn{4}{c}{Evaluation model} \\
    {} & {} 
    & DeiT-Tiny & Swin-Tiny & ConvNeXt-Tiny & ViT-Small \\
    \midrule
      & RDED               &  14.0 & 29.2 & 48.3 & - \\
    10 & EDC                & 18.4 & 38.3 & 54.4 & 14.9 \\
    \rowcolor{mygray}
      & $\text{Ours}_{10\times}$ 
                           & \textbf{23.6} & \textbf{42.7} & \textbf{55.0} & \textbf{19.8} \\
    \midrule
      & RDED               &  44.5 & 56.9 & 65.4 & - \\
    50 & EDC                & 55.0 & 63.3 & 66.6 & 54.5 \\
    \rowcolor{mygray}
      & $\text{Ours}_{10\times}$ 
                           & \textbf{56.1} & \textbf{64.3} & \textbf{66.8} & \textbf{60.7} \\
    \bottomrule
  \end{tabular}
  \end{adjustbox}
  \vspace{0.4em}
  \caption{Cross-architecture generalization on ImageNet-1K.}
  \label{tab:additional_crossarch_1K}
\end{table}

\begin{table}[!htbp]
\centering
\begin{adjustbox}{width=1.0\linewidth}

\begin{tabular}{l|cc|cc}
\toprule
Method & \multicolumn{2}{c|}{IPC = 10} & \multicolumn{2}{c}{IPC = 50} \\
\cmidrule(lr){2-3} \cmidrule(lr){4-5}
& w/o SSRS & w/ SSRS & w/o SSRS & w/ SSRS \\
\midrule
Ours$_{10\times}$ & 49.7 $\pm$ 0.2 & \textbf{50.0 $\pm$ 0.1} & 58.4 $\pm$ 0.1 & \textbf{58.9 $\pm$ 0.1} \\
Ours$_{20\times}$ & 49.4 $\pm$ 0.2 & \textbf{49.9 $\pm$ 0.1} & 58.0 $\pm$ 0.2 & \textbf{58.8 $\pm$ 0.04} \\
\bottomrule
\end{tabular}
\end{adjustbox}
\vspace{0.5em}
\caption{Ablation study on the effect of Early Smoothing–later Steep Learning Rate Schedule (SSRS) under different acceleration factors on ImageNet-1K. SSRS consistently improves performance and accelerates convergence across both IPC = 10 and IPC = 50.}
\label{tab:ssrs_ablation}
\end{table}

\noindent \textbf{Full-size Image Initialization} We adopt real image initialization sampled from the original dataset, which accelerates distillation compared to random noise, as observed in prior works~\cite{Acc-DD,dwa2024neurips, Shao_NEURIPS2024_EDC,shen2024deltsimplediversitydrivenearlylate,cui2025datasetdistillationcommitteevoting}. Our closest baseline, EDC, relies on RDED~\cite{sun2024diversity} for fast, optimization-free initialization by composing each synthetic image from several high-confidence patches of real data. 
However, patch‑based compression oversamples similar regions, which leads to redundancy, and it further distorts local features because of the compressed representation. We further posit that the subsequent relabeling stage, which functions as knowledge distillation, may be less effective under patch-based representations, as fragmented local regions provide limited global context for transferring semantic knowledge from teacher to student.
In contrast, full-image initialization preserves global semantics and spatial coherence, improving generalization and convergence while simplifying deployment by removing the pre-synthesis step.  At IPC = 1, we adopt RDED for its stronger performance under extreme data scarcity, but we downplay this case because single-image settings yield unstable accuracy across runs.

\noindent \textbf{Initialization Sensitivity in Extremely Compressed IPC}

\noindent Table~\ref{tab:appendix-ipc1-instability} reports three runs of our method at IPC = 1 with different RDED initialization seeds. Performance varies by up to 2\%, highlighting the instability caused by different initializations and the challenge of fair comparison in such an extreme compression setting.

\begin{table}[!ht]
\centering
\begin{tabular}{lccc}
\toprule
{} & Run 1 & Run 2 & Run 3 \\
\midrule
Ours$_{10\times}$ & 10.89 & 12.29 & 12.90 \\
\bottomrule
\end{tabular}
\vspace{0.5em}
\caption{Performance across three independent runs of our method with different RDED initialization seeds on ImageNet-1K at IPC = 1 using a ResNet-18 architecture.}
\label{tab:appendix-ipc1-instability}
\end{table}

\noindent \textbf{Full-size vs. Patch-Based Initialization}  
\label{ablation_initialization}

We compare our full-size image initialization with EDC's patch-based initialization. 
For IPC 10, EDC initializes each synthetic image using RDED with four crops per image setting, while our method directly samples full-size images. 
For IPC 50, to test an alternative setup and further validate our method's effectiveness, we set EDC to be initialized from a single crop per image, a setting that empirically yields better accuracy, with SSRS learning schedule applied to both methods. 
All results are reported without the optimization stage to isolate the effect of initialization. 
As shown in \cref{tab:ablation_initialization}, full‑size initialization consistently yields higher accuracy than patch‑based initialization, providing a stronger starting point that enables subsequent optimization to be more efficient.

\begin{table}[h]
\centering
\vspace{0.5em}
\setlength{\tabcolsep}{21pt}
\begin{tabular}{l|cc}
\toprule
\multirow{2}{*}{Method} & \multicolumn{2}{c}{IPC} \\

 & 10 & 50 \\
\midrule
EDC & 44.5 & 57.1 \\
Ours & \textbf{47.9} & \textbf{58.4} \\
\bottomrule
\end{tabular}
\vspace{0.5em}
\caption{\textbf{Effect of initialization strategy.} Accuracy on ImageNet‑1K when comparing the patch‑based initialization of EDC with the full‑image initialization adopted in our method (E$^2$D) across two IPC settings.}
\label{tab:ablation_initialization}
\end{table}

\noindent \textbf{Relabeling and Post-Evaluation Strategies}

For relabeling, we follow the standard state-of-the-art approach for generating soft labels from the teacher model. During student training on distilled data, we adopt the \textit{Early Smoothing–Later Steep Learning Rate Schedule} (SSRS) introduced in EDC~\cite{Shao_NEURIPS2024_EDC}. SSRS combines  the advantages of smooth cosine decay and MultiStep schedules by applying a gradual cosine reduction in the early phase, followed by a sharp decay in the final phase to accelerate convergence as described in this equation:
\begin{equation}
\mu(i) =
\begin{cases}
    \displaystyle\frac{1 + \cos\left(\frac{i\pi}{\zeta N}\right)}{2}, & i \leq \frac{5N}{6}, \\[10pt]
    \displaystyle\frac{1 + \cos\left(\frac{5\pi}{6\zeta}\right)}{2} \cdot \frac{6N - 6i}{6N}, & i > \frac{5N}{6},
\end{cases}
\label{eq:SSRS}
\end{equation}
where $\mu(i)$ is the learning rate at iteration $i$, $N$ is the total number of training iterations, and $\zeta$ controls the cosine smoothness.  
Unlike EDC, which introduced SSRS only in the appendix and considered it in high‑IPC settings, our method demonstrates that applying this schedule at moderate IPC also improves accuracy. Additionally, it reduces the synthesis time required to reach peak accuracy, highlighting a connection between student training dynamics and distillation efficiency.

\noindent \textbf{Effectiveness of Early Smoothing-later Steep Learning Rate Schedule (SSRS)}

\noindent To evaluate the effectiveness of SSRS, we compare our method with and without it under various IPC settings. While performance drops significantly at extremely low IPC values (e.g., IPC = 1), SSRS consistently improves results at higher IPCs by achieving better accuracy and requiring fewer optimization steps to reach convergence. As shown in \cref{tab:ssrs_ablation}, the benefits of SSRS become increasingly clear as IPC increases. Our method, accelerated $20\times$ with SSRS, achieves near-converged performance without additional computational cost. This observation is consistent with prior work~\cite{Shao_NEURIPS2024_EDC}, which shows that an effective learning schedule can yield substantial improvements with minimal overhead. Our findings further establish a connection between optimization efficiency and the design of the learning schedule, showing that accelerated learning scheduling can significantly reduce the need for extensive optimization.

\noindent \textbf{Synthetic Data Comparison in Logit Space}

\begin{figure}[t]
    \centering
    \includegraphics[width=\linewidth]{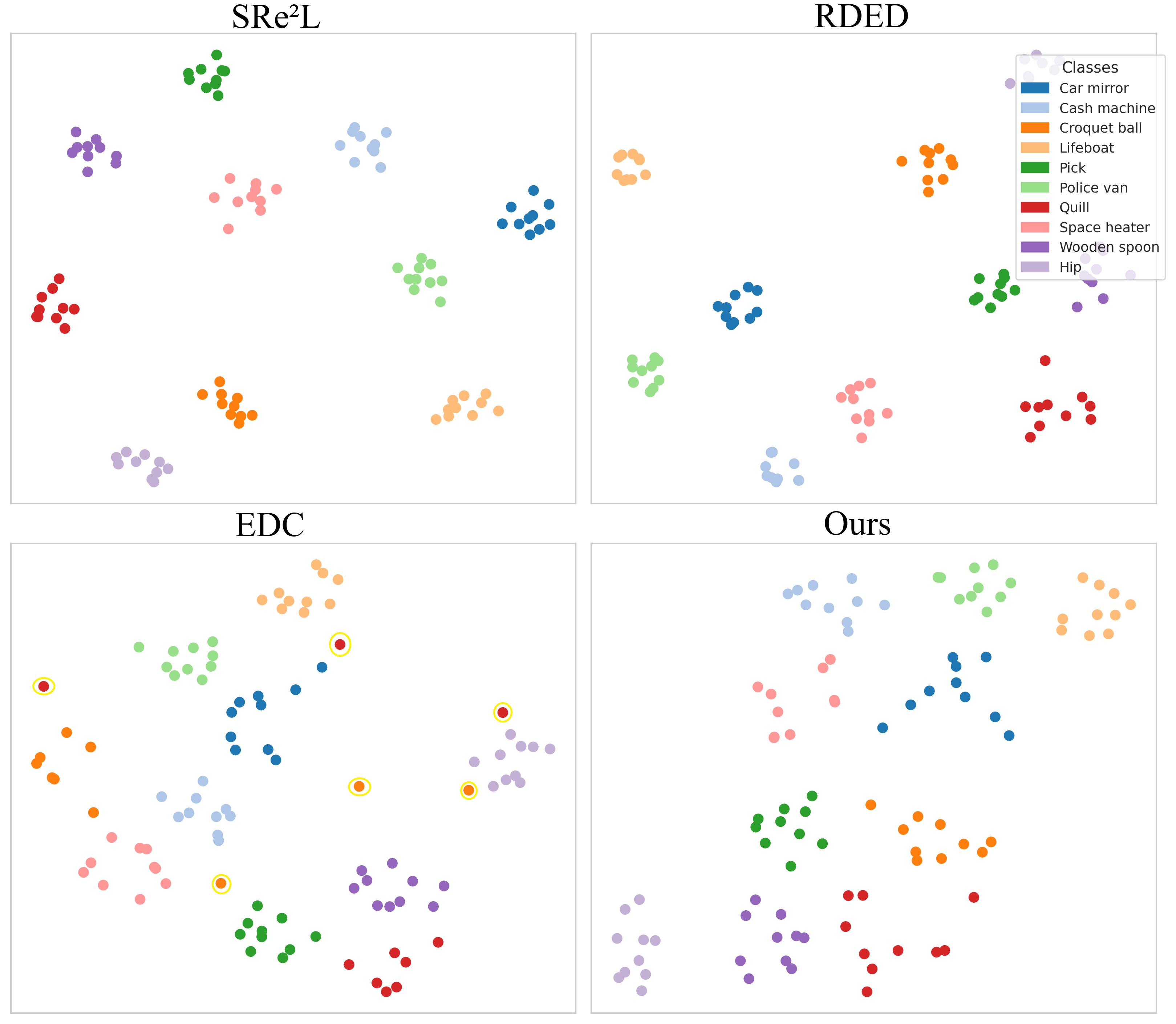}
    \vspace{0.5em}
\caption{t-SNE visualization of logit embeddings for synthetic data generated by our method and state-of-the-art baselines on ImageNet‑1K. Our method exhibits broader dispersion and clearer clusters, indicating better generalization and stronger class separability.
Outlier points outside class clusters are highlighted by yellow circles.}
\label{fig:tsne_comparison}
\end{figure}

\noindent Figure~\ref{fig:tsne_comparison} presents t-SNE visualizations of logit embeddings for synthetic data generated by our method and recent state-of-the-art baselines on ImageNet-1K.
Our synthetic representations occupy a broader region of the embedding space, indicating higher diversity and stronger generalization.
At the same time, class-specific clusters remain clearly separated, reflecting preserved discriminability and semantically coherent structure.
These observations show that our method generates synthetic data that is diverse and semantically aligned with class structure.

\noindent \textbf{Acceleration on Low-Resolution Datasets.}

\noindent We exclude low-resolution datasets such as CIFAR-10 and CIFAR-100 from our primary focus, as EDC already achieves peak performance with minimal optimization, rendering further acceleration unnecessary.
Table~\ref{tab:cifar100_accel} examines the effect of optimization iterations on CIFAR-100 using EDC. Although the default configuration uses 2000 iterations, our findings show that omitting optimization entirely preserves final accuracy for both IPC = 10 and IPC = 50 without noticeable degradation. This suggests that, for low-resolution datasets, the knowledge distillation process embedded in the relabeling stage alone effectively drives performance, making extra optimization of synthetic data redundant. Similarly, for CIFAR-10, the default EDC configuration uses only 75 optimization steps, a setting that does not benefit from further acceleration.  

\vspace{0.5em}
\begin{table}[h]
\centering
\begin{tabular}{lcc}
\toprule
Method & IPC 10 & IPC 50 \\
\midrule
EDC \textit{(reported)} & $63.7 \pm 0.3$ & $68.6 \pm 0.2$ \\
\midrule
\multicolumn{3}{l}{\textit{EDC (reproduced)}} \\
\quad No optimization & $65.1 \pm 0.2$ & $68.5 \pm 0.2$ \\
\quad With optimization & $65.2 \pm 0.1$ & $68.8 \pm 0.2$ \\
\bottomrule
\end{tabular}
\vspace{0.5em}
\caption{\textbf{Effect of optimization iterations in EDC on CIFAR-100.} The top row reports the value from the original paper; the last two rows are our reproduced results. Final accuracy remains stable even without optimization, indicating that the relabeling stage alone suffices for low-resolution datasets.}
\label{tab:cifar100_accel}
\end{table}

\section{Implementation Details}

\noindent \textbf{Loss Threshold}

\noindent The threshold $\epsilon$ specifies the minimum loss during exploration stage that a crop must reach to be selected for exploitation in our method. Conceptually, $\epsilon$ acts as a filter that identifies which regions of the synthetic image provide sufficiently strong supervisory signal to justify further optimization. The threshold controls the trade-off between optimization duration and computational overhead. Lower values increase the number of retained crops, which raises computational cost, while higher values risk premature stopping, leading to suboptimal optimization. Empirically, we found that $\epsilon = 0.5$ is effective for ImageNet-1K and $\epsilon = 0.3$ for ImageNet-21K.

\noindent \textbf{Hyperparameter Settings}

\noindent We report the hyperparameters used for each dataset experiment to facilitate reproducibility. 
\Cref{tab:hyperparams_imagenet1k} lists the configuration for ImageNet‑1K, while 
\Cref{tab:hyperparams_imagenet21k} shows the settings for ImageNet‑21K.

\section{Additional Visualization}

\begin{figure*}[t]
    \centering
    \includegraphics[width=\linewidth]{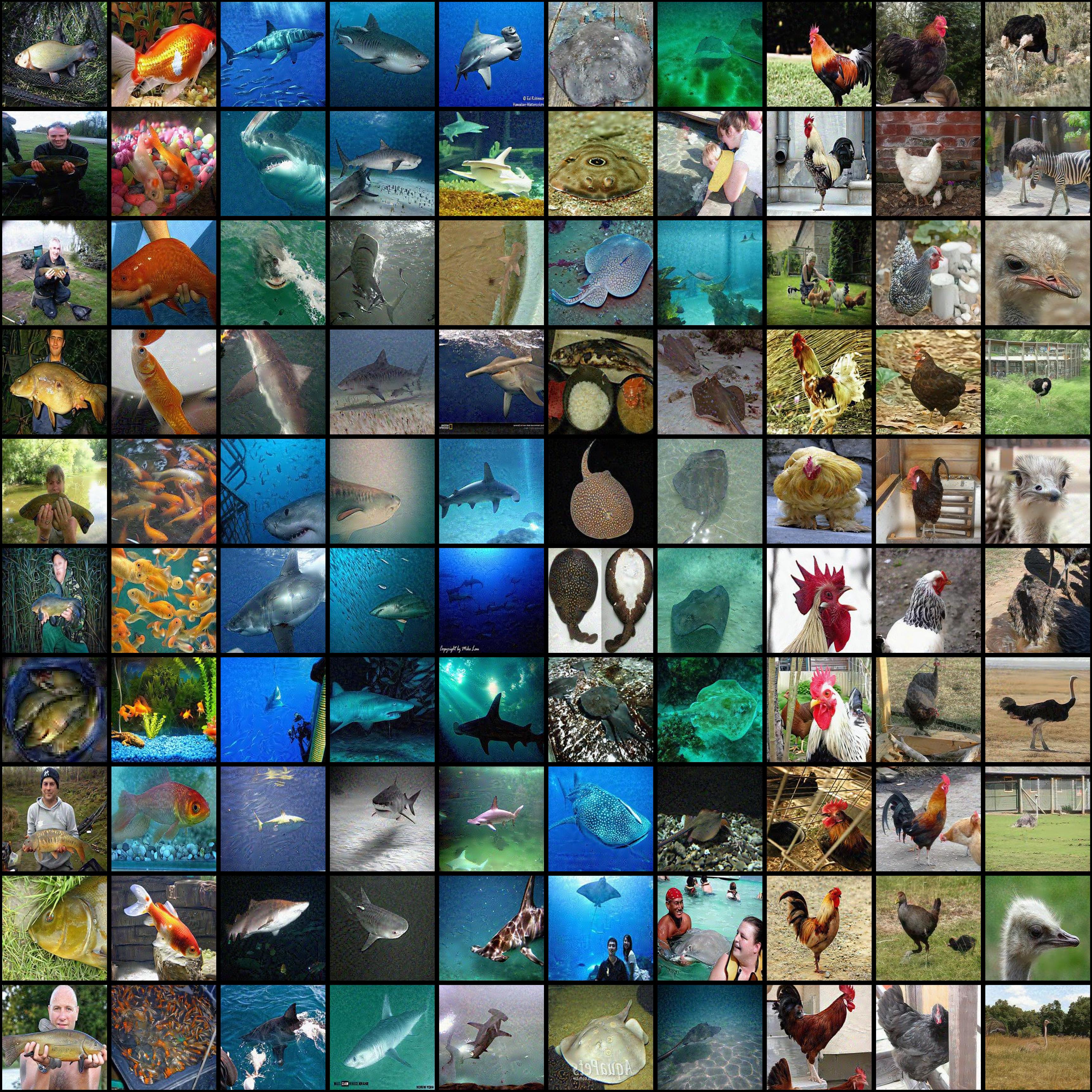}
    \vspace{0.1em}
    \caption{
        Visualization of synthetic images produced by our method on ImageNet-1K.
    }
    \label{fig:grid_ours_1k}
\end{figure*}

\begin{figure*}[t]
    \centering
    \includegraphics[width=\linewidth]{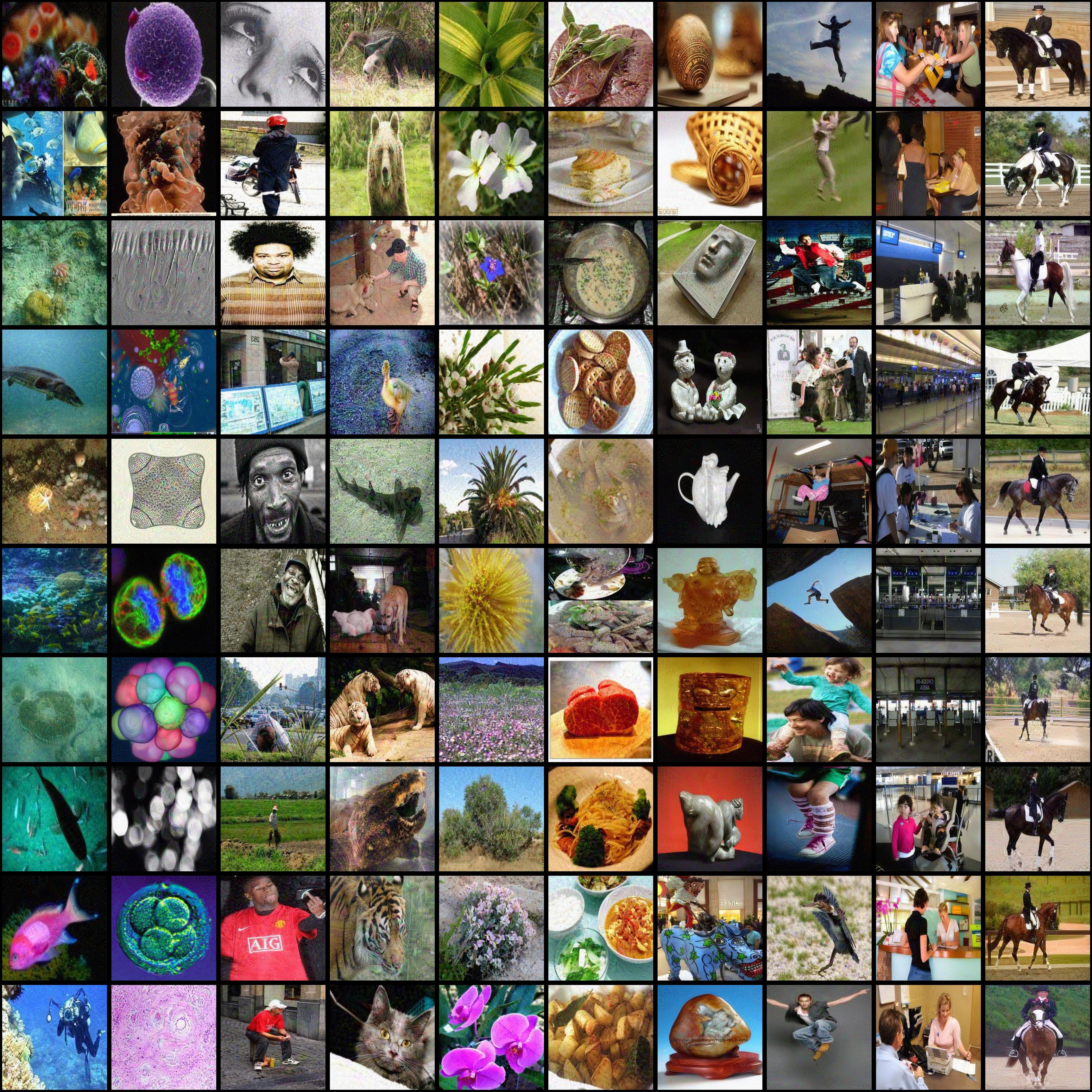}
    \vspace{0.1em} 
    \caption{
        Visualization of synthetic images produced by our method on ImageNet-21K.
    }
    \label{fig:grid_ours_21k}
\end{figure*}

We present additional visualizations of the synthetic data generated by our method for ImageNet-1K and ImageNet-21K. By relying on full-size feature representations, our method avoids the distortions arising from spatially compressed features and produces images with higher structural fidelity and clearer class-specific patterns. Illustrative examples for ImageNet-1K are shown in \cref{fig:grid_ours_1k}, while corresponding ImageNet-21K examples are provided in \cref{fig:grid_ours_21k}.

\clearpage

\clearpage

\end{document}